\documentclass{article}
\usepackage[margin=1in]{geometry}  

\usepackage[utf8]{inputenc} 
\usepackage[T1]{fontenc}    
\usepackage{hyperref}       
\usepackage{url}            
\usepackage{booktabs}       
\usepackage{amsfonts}       
\usepackage{nicefrac}       
\usepackage{microtype}      
\usepackage{xcolor}         
\usepackage{amsmath}
\usepackage{algorithm}
\usepackage{algpseudocode}
\usepackage{enumitem}
\usepackage{multirow}
\usepackage{graphicx}
\usepackage{subcaption}

\usepackage{soul}
\usepackage{bbm}
\usepackage{dsfont}
\usepackage{authblk}  
\usepackage{natbib}
\usepackage[nameinlink]{cleveref}

\title{Counterexample Guided Learning in the Large using Reasoning Agents}

\author[1]{Hongyi Liu}
\author[1]{Frederic Sala}
\author[1]{Thomas Reps}
\author[1]{Adithya Murali}

\affil[1]{Department of Computer Sciences, University of Wisconsin-Madison}

\affil[ ]{\texttt{\{hliu794, fredsala, reps, adithyamurali\}@cs.wisc.edu}}

\date{}

\begin{document}

\maketitle

\begin{abstract}
%
%

LLMs and LLM agents should improve when given feedback, but identifying when they are able to do so is difficult: 
feedback is heterogeneous, domain-specific, and difficult to control. 
We approach this challenge by asking LLMs to perform regular-expression induction, a classical symbolic learning problem where precise mechanisms for feedback exist in the form of counterexamples. 
%
%
In counterexample-guided learning, a learner (LLM) proposes candidate regular expressions from positive/negative-labeled strings, and the teacher (verifier) returns counterexamples showcasing the difference between the candidate and target languages. 
We identify novel counterexample-guided refinement strategies that enable effective regex learning, such as regularization and symbolic counterexample clusters. We also explore agentic strategies such as reflection and repair loops. 
Empirically, we find that verifier feedback substantially improves sample efficiency on challenging regex-induction tasks, reducing the number of labeled examples required and enabling learning of complex target expressions where standard prompting fails. 
For example, on the hardest task groups, our counterexample-guided framework improves success from $3.2\%$ to $38.1\%$ and from $38.9\%$ to $74.1\%$ on two different regex domains.
These results suggest that LLMs can benefit from rich feedback beyond treating it as additional data, opening the door for robust verifier-guided methods for LLM-based program synthesis and formal reasoning.

\end{abstract}

\section{Introduction}

Models, agents, and systems are increasingly required to learn and adapt from feedback.
For example, coding agents must respond to errors and unit test failures, tool-using agents must  revise their plans after failed API calls, and personalized assistants need to adapt their responses over time based on user corrections and preferences.  
Despite the importance of this capability, measuring how well models use feedback is surprisingly challenging. 
Feedback is diverse, heterogeneous, often closely tied to the task or domain, and difficult to control systematically.

Is there a simple controllable way to measure models and agents’ ability to exploit feedback? 
We make the case for \emph{regular-expression induction} as a clean testbed for measuring this capability.
Given a set of labeled strings, the goal is to infer a description of the underlying language in the form of a regular expression by generalizing from the examples. 
This task is a canonical problem in symbolic learning, and a highly challenging one: many candidate expressions can fit a small set of labeled strings, and naive in-context learning can easily overfit to superficial patterns. 
The exploration of LLM capabilities in this setting revisits a long line of work in automata learning and program synthesis 
through a modern lens.

This problem, and in fact our larger goal, are both not fully addressed by prior art. 
Traditional neural sequence models such as RNNs can treat regex recognition as a classification problem, but they do not naturally expose symbolic hypotheses and do not clearly benefit from counterexample-only learning. Conversely, LLMs can generate explicit regexes, but current usage is often prompt-driven and heuristic, with limited structure for how verifier feedback should be incorporated across rounds. Prior works also largely focus on synthesizing regexes from natural-language descriptions, as opposed to pure example-based learning~\cite{TANG2026103762,siddiq2024regexeval}. On the other hand, while counterexample-guided methods are well established in symbolic synthesis~\cite{Solar-Lezama2013,Abate2018Counterexample,Alur2013SyntaxGuided}, it remains unclear how to adapt them to modern LLM-based inference in a way that is sample-efficient, stable, and robust to overfitting. Furthermore, traditional program-synthesis techniques can stall as the number of examples becomes large. Our work contributes a much needed synergy between the two kinds of approaches, framing regex induction as an oracle-guided symbolic generation problem where the model iteratively refines its hypothesis through verifier feedback in the form of (potentially a large number of) counterexamples.

\begin{figure}[t]
    \centering
    \includegraphics[width=\linewidth]{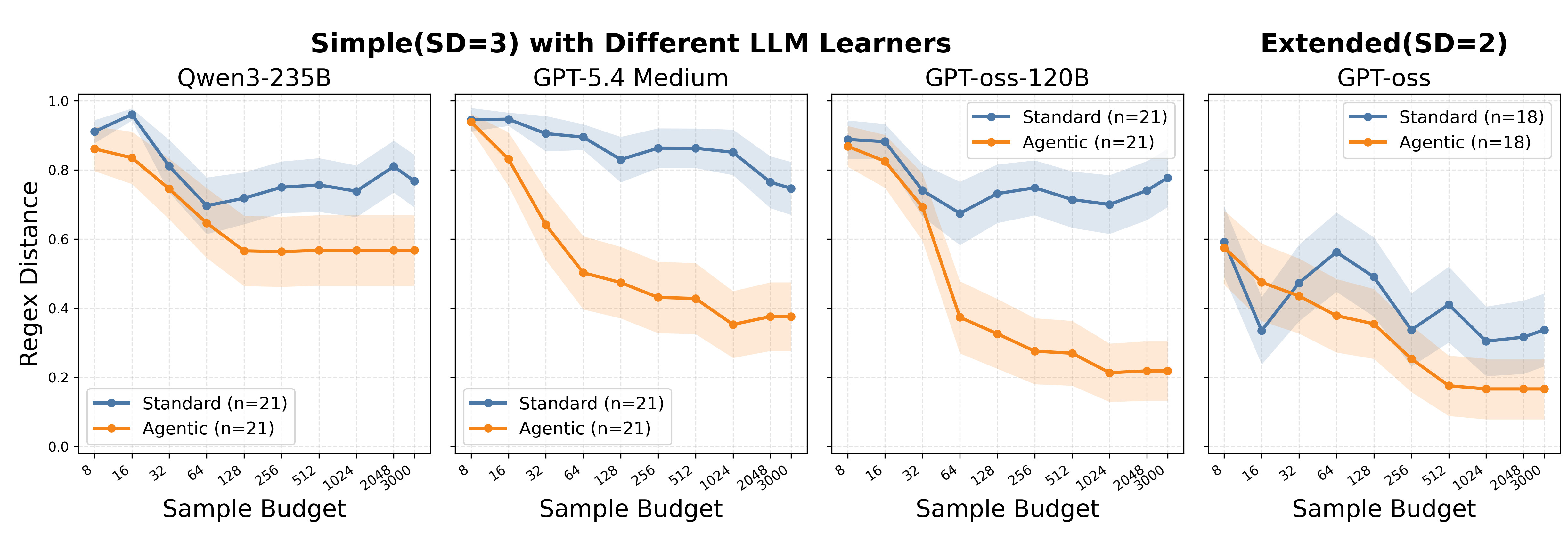}
    \caption{Learning dynamics of our agentic counterexample-guided learner (orange) versus standard prompting without counterexamples (blue). As the number of examples increases, the hypothesis returned by the agentic learner converges closer\protect\footnotemark and faster to the target concept, showing the value of leveraging rich feedback. Points show the mean distance and shaded regions show variation.
    } 
    
    \label{fig:acc_eval_sample_budget_comprehensive}
    \vspace{-1em}
\end{figure}

Concretely, in this work we present a counterexample-guided framework for LLM-based regex learning. Our approach uses LLMs and agents to generate candidate regexes from labeled strings, checks them against a target expression, and synthesizes informative counterexamples 
from the symmetric difference between the predicted and target languages. We further study (a) regularization strategies that bias the model toward shorter, simpler, and more plausible symbolic hypotheses, (b) richer variants of counterexamples that cluster several individual counterexamples with common failure information, and (c) agentic strategies 
that use reflection to extract the structural implications of counterexamples and repair loops to iteratively revise failed regex hypotheses.
Empirically, we find that counterexample-guided learning is especially helpful in the symbolic generation setting: on several challenging regexes, it substantially reduces the number of training examples required and, in some cases, enables successful recovery where standard prompting fails, as shown in Figure~\ref{fig:acc_eval_sample_budget_comprehensive}.

\noindent \textbf{Our Contributions.} The main contributions of this paper are: 

\begin{itemize}[leftmargin=*,topsep=0pt, noitemsep]
\item \textbf{An oracle-guided formulation of LLM-based regex induction:} We formulate regex induction with LLMs as an iterative symbolic learning problem in which candidate regexes are refined using feedback from an oracle in the form of counterexamples.

\item \textbf{An active learning methodology for regex induction using LLMs:}
To build effective LLM-based active learners for regex induction, we lift the classic teacher-learner learning framework to use LLM-based learners, investigating techniques such as regularization to prioritize simpler hypotheses and to cluster counterexamples symbolically.

\item \textbf{An agentic workflow for counterexample-guided learning in the large:} We build novel agentic frameworks that learn regular expressions by reasoning about counterexamples across many rounds of learning, using techniques such as agentic reflection and iterative expression repair.

\item \textbf{A rigorous evaluation of the dynamics of LLM learning with verifier feedback:} We study the performance of several counterexample-based learners across two datasets for regex induction. We show that our agentic framework improves learning performance compared to standard prompting from $3.2\%$ to $38.1\%$ and $38.9\%$ to $74.1\%$ on the hardest instances in the two benchmark suites. We further show that our technique improves sample efficiency and that these gains persist across various base model families. We also compare our technique against baselines and perform ablation studies, demonstrating that each component of our agentic framework contributes significantly.

\end{itemize}

\section{Related Work}

\noindent\textbf{Regular-Language Learning or Regex Induction.}
Learning regular languages from examples and queries has a long history in the study of formal languages. A classic example is Angluin's $L^\ast$ algorithm~\cite{Angluin1987}, which learns regular sets through membership and equivalence queries with counterexamples; this oracle setting closely matches ours. Beyond query-based learning, prior work has also studied direct regex inference from examples, including learning from positive data~\cite{Fernau2009}, search-based synthesis from labeled examples~\cite{Lee2016Synthesizing,Bartoli2014Automatic}, and constraint- or optimization-based approaches~\cite{Gao2020IntegerProgramming}. Prior to LLMs, neural methods also explored regular-language learning with recurrent models trained on labeled strings, sometimes followed by automata extraction or analysis of learned finite-state structure~\cite{Cohen2017Inducing,Oliva2021Stability}. These approaches provide important background, but they do not study regex generation with verifier-guided refinement in an LLM-centered setting.

\noindent\textbf{Counterexample-Guided Symbolic Synthesis.}
Counterexample-guided refinement is also central to symbolic synthesis. Counterexample-Guided Inductive Synthesis (CEGIS) combines candidate generation with automated validation and counterexample generation, and has been highly successful in program synthesis and related symbolic-reasoning tasks~\cite{Solar-Lezama2013,Abate2018Counterexample}. Beyond program generation, counterexample-based supervision has also been used in symbolic learning problems, such as invariant synthesis~\cite{Garg2014ICE,Garg2016DecisionTrees} and automata-related reasoning~\cite{Weiss2018AutomataExtraction}, where verifier feedback helps expose mistakes that are difficult to identify from positive and negative examples alone. Our work is aligned with this line of research, but adapts it to an LLM-centered setting in which the learner is not a symbolic search procedure, but a language model that must interpret counterexamples and revise its own symbolic hypothesis.

\noindent\textbf{LLMs for Symbolic Reasoning, Program Synthesis, and Regex Induction.}
Recent work has shown that LLMs can support symbolic reasoning and program synthesis by generating structured outputs such as code, logical forms, and symbolic rules~\cite{Chen2021Codex,Austin2021ProgramSynthesis,Shin2021ConstrainedLM}. In regex-related tasks, prior work has shown that language models can generate explicit regular expressions from natural-language descriptions or prompts, often combined with examples, repair heuristics, or multimodal and sketch-based synthesis mechanisms rather than pure induction from labeled strings alone~\cite{TANG2026103762,siddiq2024regexeval,Chen2020Regel,Ye2020SketchDriven}. Furthermore, LLMs are increasingly used in settings with external feedback, including process-supervision methods that verify intermediate reasoning steps~\cite{Lightman2023LetsVerify} and reinforcement learning with verifiable rewards~\cite{Liu2025TrustButVerify}. Our work differs from these directions by studying LLM-based regex induction from labeled strings, and by asking not only whether verifier-provided feedback helps, but also how it can be incorporated.

\footnotetext{Distance is measured using Diff Ratio, which is the fraction of strings up to a fixed length ($32$ in the figure) on which the hypothesis regex and target regex disagree, normalized by the strings accepted by either regex.}

\section{Problem Statement and Motivation}

Our study is motivated by the long history of work on active learning where a learner actively solicits feedback from a teacher, and uses the feedback to `tease out' the hypothesis over several rounds of interaction. Specifically, in the realm of symbolic learning/reasoning tasks, the capability frontier of LLMs can be spiky, and there is little consensus on broad approaches for harnessing these abilities well. The setting of oracle-aware learning through iterative feedback offers a richer framework to elicit symbolic learning capabilities from LLMs. Simultaneously, traditional search-based approaches for counterexample-guided learning can suffer when challenging problem instances accrue many counterexamples over many rounds, and machine-learning-based approaches provide a compelling alternative to address this issue. Together, these facts motivate the high-level research question of this paper: what can LLMs and Agents bring to counterexample-guided learning in the large?

\noindent \textbf{Oracle-Guided Learning.} We formulate the problem we study as an active-learning problem guided by an oracle/teacher. 
Let $\mathcal{X}$ be a domain of elements. We refer to subsets of this domain as \emph{languages}. Let $L^\star \subseteq \mathcal{X}$ denote the target language we wish to learn. The learner observes finite evidence about the target language in the form of labeled elements from the domain, and synthesizes a candidate language. 
The teacher then inspects the proposed candidate and provides feedback in the form of labeled elements that witness the incorrectness of the proposed candidate, namely elements that should be included in the target language but are not included in the candidate language, and vice versa. Note that in a concrete realization of this framework, the teacher does not necessarily need access to the target concept to provide such feedback: it can analyze the candidate with respect to test cases, constraints, or other properties it knows to be true about the target, and return witnesses that showcase violations of these properties.


\noindent \textbf{Problem Statement: Regular-Language Induction.} We instantiate the above framework for regular-language induction. The instance space $\mathcal{X}$ is a set of finite strings $\Sigma^\star$, where $\Sigma$ is a finite alphabet. 
The target languages we study are those that 
can be represented by a regular expression or, equivalently, a finite-state automaton. Given a regular expression $r$, we denote the language it represents by $L(r)$. Given labeled examples, a learner must synthesize a regular expression $\hat r$ such that $L(\hat r)$ matches the target language $L^\star$. This task is both predictive and symbolic: the output must classify strings correctly, but it must also be an interpretable expression that can be compiled, verified, and compared against the target. We construct the teacher by sampling from the symmetric difference $L(\hat r)\triangle L^\star$. We can build such a teacher in this setting because regular languages are closed under Boolean operations, and there exist efficient off-the-shelf libraries for constructing a symmetric-difference automaton~\cite{pyformlang}.


\section{Reasoning Agents for Counterexample-Guided Inductive Learning}
\label{sec:methodology}

\begin{figure}[t]
    \centering
    \includegraphics[width=0.85\linewidth]{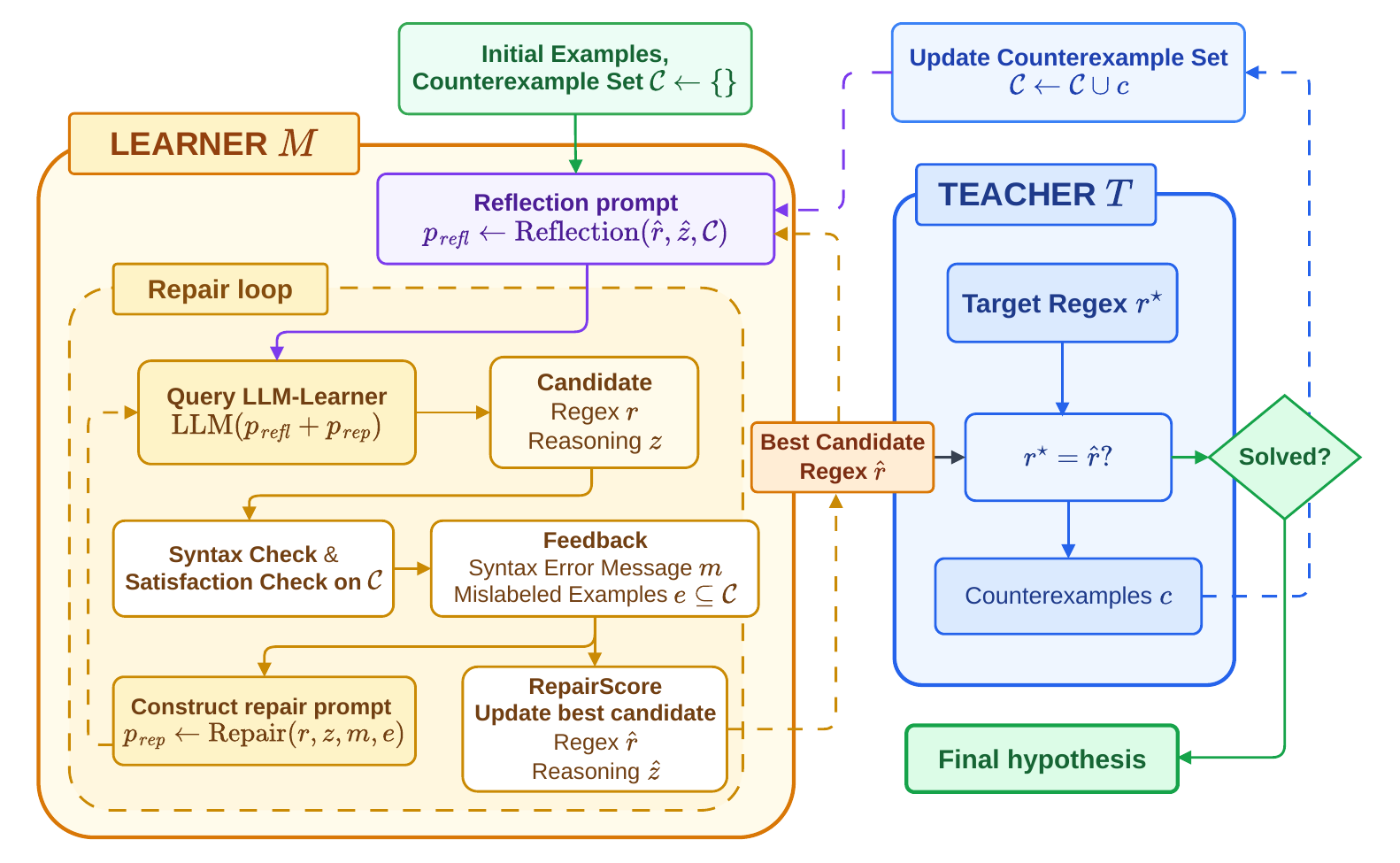}
    \caption{Overview of our iterative counterexample-guided agentic refinement framework with a teacher and a learner. The learner uses agentic reflection and repair along with counterexamples (see Section~\ref{sec:methodology}). The teacher constructs counterexamples by checking equivalence between the candidate and target regexes. 
    }
    \label{fig:overview}
    \vspace{-1em}
\end{figure}

We describe the various LLM-based learners that we built for counterexample-guided learning. Figure~\ref{fig:overview} describes the overall construction. We first present the teacher-learner interaction, and then describe the key parts of our learning framework. We provide full prompt templates in Appendix~\ref{appendix:prompts}.

\noindent \textbf{Teacher-Learner Iterative Framework.} 
The learning framework is built around the teacher-learner setting described earlier. The learning proceeds in several \emph{rounds}. In each round, the learner proposes a candidate regular expression. The teacher analyzes it, and determines if it is equivalent to the target language. Note that the equivalence of regular languages is decidable, and can be performed efficiently by existing libraries. 

If the proposed candidate is not equivalent, the teacher returns both positive and negative counterexamples, as applicable. The learner adds these counterexamples to the existing set (provided as part of the prompt) and attempts to synthesize a better candidate. The process continues until the learner finds the target or reaches a predetermined maximum attempt bound. 
%
Note that this process is a closed-loop refinement procedure. Unlike learning paradigms where the training set is fixed in advance, supervision here is adaptive: the teacher selects feedback based on the learner's current errors. We argue that this approach is especially useful for symbolic synthesis, where a hypothesis may fit a limited sample dataset but may fail on unseen instances.

\noindent \textbf{Improving Effectiveness of Counterexamples.} 
\label{method:counterexamples}
We synthesize counterexamples as follows: given the candidate regular expression $r$ and the target $r^\star$, we construct the corresponding minimal finite automata $A$ and $A^\star$. Positive counterexamples (i.e., strings that must be included in the target language but are not a member of the candidate language) are sampled from the difference $A^\star \setminus A$, and negative counterexamples are sampled from $A \setminus A^\star$. The sampling procedure traverses the structure of the difference automata and returns a diverse set of strings that cover different paths, representing a varied set of ``mistakes'' in the proposed concept. 


However, the number of paths in an automaton can grow rapidly. 
Multiple symbols may induce the same state transition, and loops---corresponding to Kleene-stars in the regular expression---generate infinitely many counterexamples. We therefore propose the idea of \textbf{clustered} counterexamples, which are symbolic counterexamples that capture many similar individual counterexample instances. 

\noindent\textit{Clustered Counterexamples.} We construct clustered counterexamples as follows. 
For a state $q$ in the minimized difference automaton, we group outgoing symbols by their destination state
\[
\Gamma(q,q') = \{ a \in \Sigma : \delta(q,a)=q' \}.
\]
Symbols in the same group are behaviorally equivalent at state $q$, because they induce the same transition. We represent each group with a new super-character symbol that we add to the alphabet. 
For example, we represent the set \{A,\ldots,Z,a,\ldots,z\} as a new super-character [A\text{-}Za\text{-}z]. We may also use negations to represent the class when convenient, e.g. $[\hat{\ }0\text{-}9]$. Singleton groups are of course identified with their literal characters.

Formally, a clustered counterexample is a path from the start state of the difference DFA to an accepting state, where each edge contributes one super-character. A clustered counterexample compactly represents many concrete strings that follow the same DFA trajectory and share the same disagreement label. For example, \textit{S[0-9]\#run} represents strings such as \textit{S2\#run} and \textit{S8\#run}. Clustered counterexamples are interpreted universally during learning: a clustered example is accepted only if every string represented by it is accepted. 
We enumerate loop-free accepting paths and use a finite frontier and expansion budget to keep counterexample generation bounded.

\noindent \textbf{Agentic Workflow.} 
We build on the above ideas by lifting the teacher-learner framework into an agentic workflow. Recent work has shown that LLMs can benefit from agentic workflows where they iteratively evaluate and improve their own hypotheses~\citep{madaan2023selfrefineiterativerefinementselffeedback,shinn2023reflexionlanguageagentsverbal,xiong2024watchstepllmagent,song2025progcoprogramhelpsselfcorrection}. We develop such an iterative agentic approach to help LLMs use counterexamples more effectively.
Our approach has an outer loop and an inner loop (see Figure~\ref{fig:overview}). The outer loop accumulates feedback from the teacher and uses \textbf{reflection} to augment the native reasoning capabilities of the underlying model. The inner loop refines the process of proposing candidate regular expressions by performing \textbf{repair} over a smaller set of counterexamples chosen from the full set of counterexamples available to the learner. The detailed algorithm is in Appendix~\ref{appendix:pseudo_code}.


We now describe the reflection mechanism and the repair loop in detail. 

\noindent\textit{Reflection.} The key idea behind reflection is to ask the LLM to identify specific conceptual errors in previous hypotheses and use them to improve future proposals. 
We therefore instruct the LLM to respond not only with a candidate regular expression, but also a short reasoning trace explaining how the synthesized expression was chosen. We then provide this reasoning trace back to the model along with counterexamples generated by the teacher. The learner is then instructed to understand the previous attempt, explain what failed, and revise the hypothesis according to the teacher's feedback. 
Concretely, we construct a reflection prompt in each epoch of the outer loop consisting of the reflection instruction, the previous hypothesis, the reasoning trace corresponding to the previous hypothesis, and the new set of counterexamples provided by the teacher.



\noindent\textit{Repair Loop.} The mere presence of examples in the context does not ensure that LLMs will return candidates that are consistent with all of them. In fact, we observe that LLMs may return candidates that are not consistent with provided instructions, are syntactically invalid, or most importantly fail to fit the accumulated training examples. However, the hypotheses they return are not completely wrong or irrelevant either. Therefore, we introduce a repair loop that repeatedly queries the LLM for candidates, providing pointed feedback on the failures. 
Specifically, we require that the following conditions are satisfied:
\begin{itemize}[leftmargin=*,topsep=0pt, noitemsep]
    \item Syntax Check: the expression must be syntactically valid;
    \item Satisfaction Check: the regex must correctly label all counterexamples seen so far ($\mathcal{C}$). 
\end{itemize}

We retry repair by providing the feedback from all previous tries until a candidate satisfying both requirements is synthesized, or a predetermined retry budget is exhausted. We score the candidates obtained in this manner as follows:
\begin{align*}
\textsc{RepairScore}(m,e,\mathcal{C})
=
\frac{1}{2}\mathrm{Compiles}(m)
+
\frac{1}{2}
\left(
1 - \frac{|e|}{|\mathcal{C}|}
\right),
\end{align*}
where $m$ is syntax error message, $\mathrm{Compiles}(m)\in\{0,1\}$ indicates whether the regex compiles, $e$ is the set of misclassified examples, and $\mathcal{C}$ is the accumulated counterexample set. The repair loop returns the highest scoring candidate  synthesized across retries.

\noindent \textbf{Regularization.} 
In preliminary investigations, we observed that LLMs often fit the given examples with 
overly specific or unnecessarily complex regular expressions. To address this issue we add prompt-level regularization:
we instruct the learner to prefer simpler hypotheses while remaining consistent with the examples. During inference, we limit the maximum length of the expression, bound the nesting depth of complex operators, and restrict the alphabet to valid symbols for the task. 


\section{Experiments}
\label{experiments}

We study the following research questions:
\begin{itemize}[leftmargin=*,topsep=0pt, noitemsep]
    \item \textbf{RQ1: Can feedback be used by models and agents, and what affects its usefulness?} We ask whether counterexample feedback improves regex learning relative to learning from randomly sampled examples alone. We study how this effect changes with task complexity.
    \item \textbf{RQ2: Can agentic setups make better use of feedback, and at what cost?} We compare single-shot approaches with agentic workflows that perform reflection and repair, studying whether additional reasoning effort improves the use of feedback, and whether those gains justify the increased token cost.
    \item \textbf{RQ3: How does the choice of backbone model affect feedback-guided induction?} We evaluate whether the benefits of feedback-guided induction persist across different model families and reasoning configurations. 
    \item \textbf{RQ4: How robust is counterexample-guided learning to prompt design?} We vary the prompt components used to specify the input \& output formats, reasoning style, and strategy. 
    \item \textbf{RQ5: How do different ways of providing counterexamples affect performance?} We compare clustered and non-clustered counterexamples to test whether feedback quality matters beyond the number of examples provided.
\end{itemize}
We also perform ablations to gauge the importance of each component in our pipeline. 

\noindent \textbf{Data.} We evaluate our methods mainly on two datasets with different regular-expression variants. A summary of the syntax for each variant and how we collect the datasets are provided in Appendix~\ref{appendix:comp-simple-extended}.

\begin{itemize}[leftmargin=*,topsep=0pt, noitemsep]
\item \textbf{Simple Regex.} The simple regex language uses the standard textbook regular-expression operators---concatenation, union, Kleene star, and epsilon---over a fixed three-symbol alphabet \(\{a,b,c\}\).
\item \textbf{Extended Regex.} The extended regular language is closer to real-world usage: beyond the simple setting, it additionally supports one-or-more, optional items, counted repetition, intersection, and negation, yielding a richer and more flexible syntax. The extended regexes we use are adapted from KB13~\cite{kushman-barzilay-2013-using}.
\end{itemize}

We measure regex complexity using two criteria: the number of states in the minimized DFA (\emph{\#States}) and the maximum Kleene-star depth (\emph{StarDepth}). We select three representative languages for each $(\emph{\#States}, \emph{StarDepth})$ combination, with \emph{\#States} ranging from $3$ to $9$, \emph{StarDepth} from $0$ to $4$ for Simple Regex, and from $0$ to $2$ for Extended Regex due to limitations of the original KB13 dataset.

\noindent \textbf{Setup.} Detailed configuration of LLM and hyperparameters can be found in Appendix~\ref{appendix:setup-details}.

\subsection{RQ1: Can feedback be used by models and agents, and what affects its usefulness?}

\noindent \textbf{Setup.} We evaluate three settings. \textsc{Standard} uses no counterexamples; instead, it constructs the training set with randomly sampled labeled strings at logarithmically increasing scales, together with regularization. \textsc{Single} uses clustered counterexamples for training data and applies regularization, but excludes the agentic workflow. \textsc{Agentic} uses the full proposed framework. All counterexamples are clustered as described in Section~\ref{method:counterexamples}.

\noindent \textbf{Results.}
\begin{table*}[t]
\centering
\small
\setlength{\tabcolsep}{5pt}

\begin{minipage}[t]{0.56\textwidth}
\centering
\begin{tabular}{lccccc}
\toprule
\multicolumn{6}{c}{Simple Regex Run-Level Success Rate$\uparrow$} \\
\cmidrule(lr){1-6}
Method & \shortstack{SD=0} & \shortstack{SD=1} & \shortstack{SD=2} & \shortstack{SD=3} & \shortstack{SD=4} \\
\midrule
Standard & 100.0\% & 50.8\% & 23.8\% & 27.0\% & 3.2\% \\
Single & 100.0\% & 92.1\% & 73.0\% & 61.9\% & 23.8\% \\
Agentic & 100.0\% & 100.0\% & 82.5\% & 77.8\% & 38.1\% \\
\bottomrule
\end{tabular}
\end{minipage}
\hfill
\begin{minipage}[t]{0.38\textwidth}
\centering
\begin{tabular}{lccc}
\toprule
\multicolumn{4}{c}{Extended Regex Run-Level Success Rate$\uparrow$} \\
\cmidrule(lr){1-4}
Method & \shortstack{SD=0} & \shortstack{SD=1} & \shortstack{SD=2} \\
\midrule
Standard & 100.0\% & 66.7\% & 38.9\% \\
Single & 100.0\% & 77.8\% & 68.5\% \\
Agentic & 100.0\% & 84.1\% & 74.1\% \\
\bottomrule
\end{tabular}
\end{minipage}

\caption{Run-level success rate by regex complexity (aggregated by \emph{StarDepth} (SD)) on the Simple Regex and Extended Regex. Each cell reports success ratio over 21 regexes and 3 runs per regex.}
\label{tab:main-result-success-by-stardepth}
\end{table*}
Table~\ref{tab:main-result-success-by-stardepth} shows that feedback in the form of counterexamples substantially improve regex induction on both datasets. On Simple Regex, \textsc{Standard} falls sharply as \emph{StarDepth} increases, reaching only $3.2\%$ success at \emph{StarDepth 4}, while \textsc{Single} and \textsc{Agentic} perform much better at $23.8\%$ and $38.1\%$, respectively. The same pattern appears on Extended Regex: at \emph{StarDepth 2}, counterexample-guided methods are better than \textsc{Standard} by nearly 30 percentage points, with \textsc{Agentic} achieving the best result. 
The benefit of counterexamples becomes more significant as tasks become harder. As \emph{StarDepth} increases, \textsc{Standard} prompting degrades quickly, while counterexample-guided settings remain substantially more reliable. This result suggests that counterexamples are most useful when the initial labeled examples underdetermine the target language and the model must recover more complex symbolic structure.

\subsection{RQ2: Can agentic setups make better use of feedback, and at what cost?}
\noindent \textbf{Setup.}
To isolate the value of the agentic workflow, we compare \textsc{Single} and \textsc{Agentic}. Both settings use clustered counterexamples and regularization, but only \textsc{Agentic} performs additional reflection and repair steps. We measure both final success rate and inference cost, where cost is approximated by total token usage.

\noindent \textbf{Results.}
The \textsc{Agentic} workflow makes more effective use of counterexamples. Across both datasets and all nontrivial \emph{StarDepth} levels, \textsc{Agentic} consistently outperforms \textsc{Single}. This observation indicates that reflection and repair loops help the model extract useful information from counterexamples and translate it into improved regex hypotheses.

\begin{figure}[t]
    \centering

    \begin{subfigure}[t]{0.48\textwidth}
        \centering
        \includegraphics[width=\linewidth]{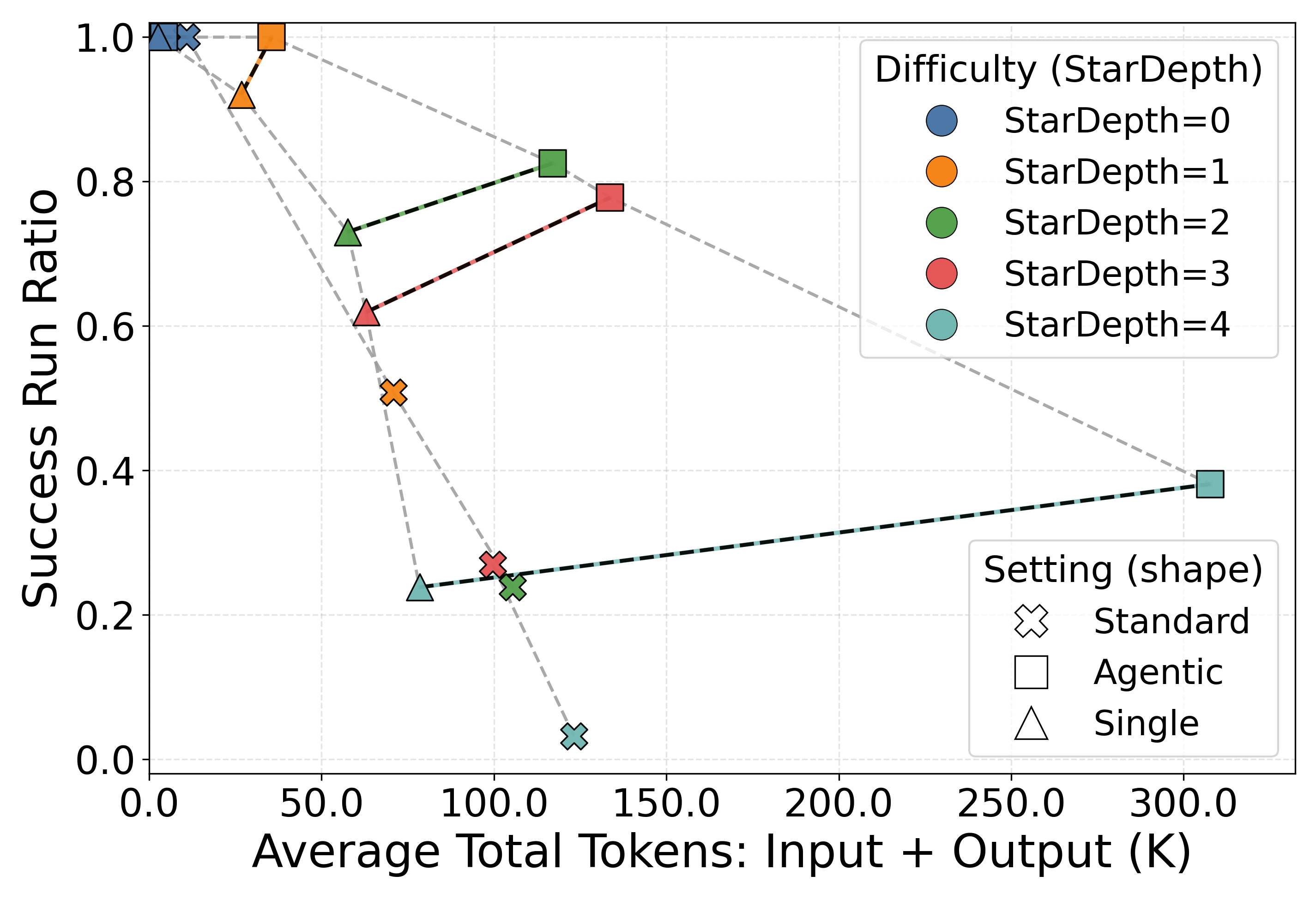}
        \caption{Main Simple Regex Dataset}
        \label{fig:pareto-simplyrx-selected}
    \end{subfigure}
    \hfill
    \begin{subfigure}[t]{0.48\textwidth}
        \centering
        \includegraphics[width=\linewidth]{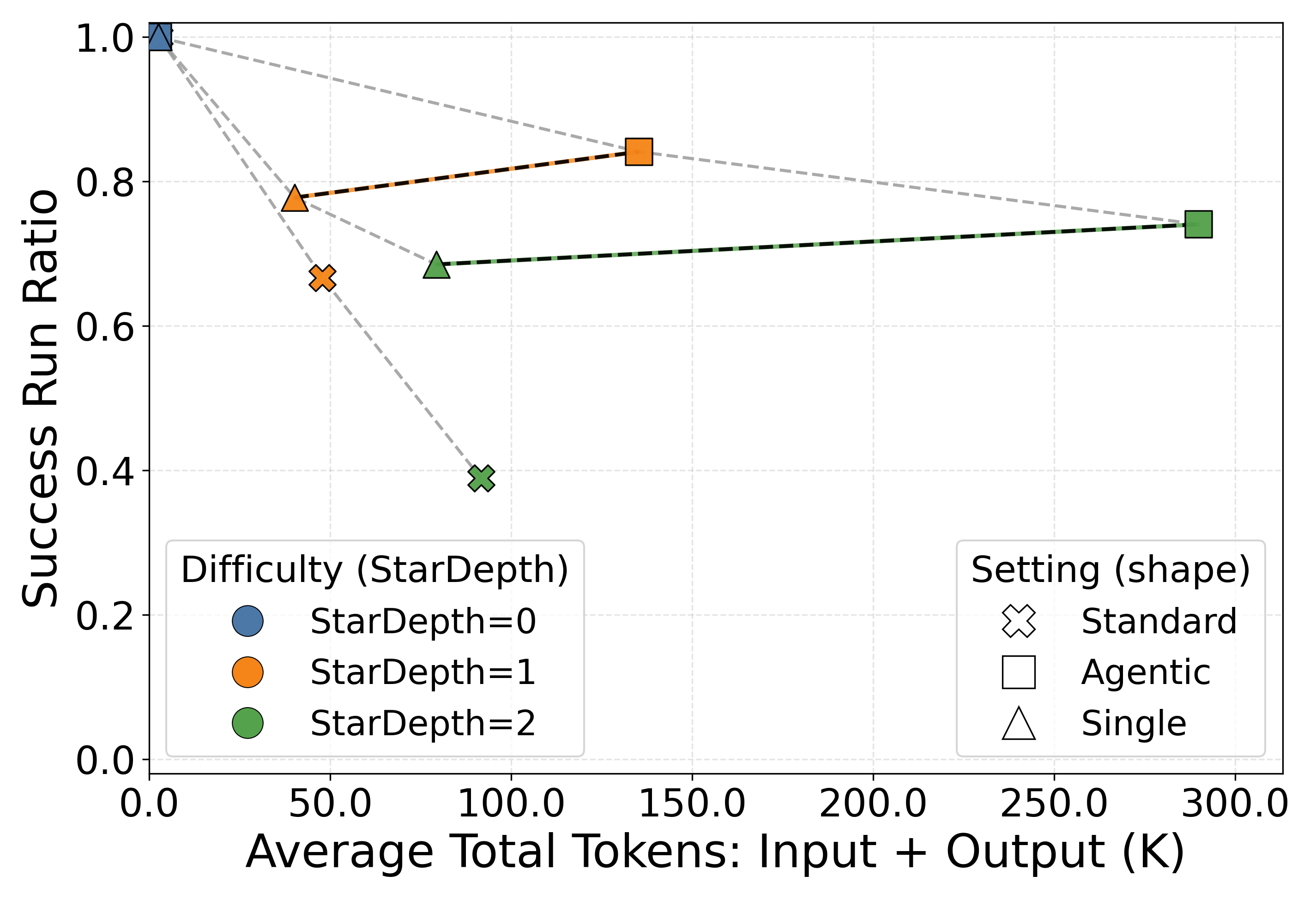}
        \caption{Main Extended Regex Dataset}
        \label{fig:pareto-extrx-selected}
    \end{subfigure}

    \caption{Pareto frontiers of the main results on Simple and Extended Regex. Points at the same \emph{StarDepth} share the same color, while marker shapes distinguish learning settings. Gray dashed lines connect the same setting across difficulty levels, showing how the cost--accuracy trade-off shifts with task difficulty. Colored solid segments with black dashed overlays indicate Pareto-optimal trade-offs.}
    \label{fig:pareto-frontiers}
\end{figure}

The Pareto plots in Figures~\ref{fig:pareto-simplyrx-selected} and~\ref{fig:pareto-extrx-selected} further show that counterexample-guided learning is more cost-efficient than the \textsc{Standard} baseline. For most nontrivial \emph{StarDepth} levels, \textsc{Single} achieves higher success rates with fewer total tokens than \textsc{Standard}. The plots also indicate the cost of the \textsc{Agentic} workflow: it typically achieves the highest success rate, but requires more tokens due to additional reflection and repair rounds. Counterexamples improve both success and sample efficiency, while the full agentic workflow further increases success but uses more tokens.

\subsection{RQ3: How does the choice of backbone model affect feedback-guided induction?}

\noindent \textbf{Setup.}
We compare different backbone models on Simple Regex while holding the task, evaluation protocol, and maximum output-token budget fixed. We evaluate \textsc{Standard} and \textsc{Agentic} across Qwen3-235B~\cite{qwen3technicalreport}, GPT-5.4 Medium, GPT-5.4 xHigh~\cite{OpenAI2026GPT54}, and GPT-oss-120B~\cite{openai2025gptoss120bgptoss20bmodel}.

\noindent \textbf{Results.} Table~\ref{tab:simplyrx-model-method-stardepth} shows that for every tested model and every nontrivial \emph{StarDepth} level, \textsc{Agentic} consistently improves over \textsc{Standard}. This result indicates that the proposed framework can broadly use counterexamples to improve regular-language induction, rather than depending on a single backbone model. However, the absolute performance still varies across models. GPT-5.4 xHigh performs best overall, while Qwen3-235B performs worst, with the other two models in between. 

Because GPT-5.4 xHigh achieves the best overall performance, we further explored its behavior on the harder \emph{StarDepth=4} split. Under the agentic setting, GPT-5.4 xHigh substantially outperforms GPT-oss-120B, achieving $85.7\%$ success, compared with $38.1\%$ for GPT-oss-120B.


\begin{table*}[t]
\centering
\small
\setlength{\tabcolsep}{6pt}
\begin{tabular}{lcccccc}
\toprule
\multirow{2}{*}{Model}
& \multicolumn{2}{c}{SD=1} & \multicolumn{2}{c}{SD=2} & \multicolumn{2}{c}{SD=3} \\
\cmidrule(lr){2-3}
\cmidrule(lr){4-5}
\cmidrule(lr){6-7}
& Standard & Agentic & Standard & Agentic & Standard & Agentic \\
\midrule
Qwen3-235B & 28.6\% & 61.9\% & 14.3\% & 38.1\% & 14.3\% & 38.1\% \\
GPT-5.4 Medium & 38.1\% & 90.5\% & 14.3\% & 47.6\% & 14.3\% & 57.1\% \\
GPT-oss-120B & 52.4\% & 100.0\% & 14.3\% & 85.7\% & 19.0\% & 76.2\% \\
GPT-5.4 xHigh & 90.5\% & 100.0\% & 71.4\% & 95.2\% & 75.0\% & 85.7\% \\
\bottomrule
\end{tabular}
\caption{Performance of different backbone models on Simply Regex aggregated by \emph{StarDepth}. Each cell reports run-level success ratio over 21 regexes (3 per \emph{\#States}), one run per regex. Qwen3-235B denotes \texttt{Qwen/Qwen3-235B-A22B-Instruct-2507-tput}. GPT-5.4 Medium and GPT-5.4 xHigh both use \texttt{openai/gpt-5.4}, with \texttt{reasoning\_effort} set to \texttt{medium} and \texttt{xhigh}, respectively.}
\label{tab:simplyrx-model-method-stardepth}
\end{table*}

\subsection{RQ4: How robust is counterexample-guided learning to prompt design?}
\noindent \textbf{Setup.} We ablate the prompt design on Simple Regex. All prompt variants include a short task instruction that asks the model to infer a regular language from labeled strings. We then vary four additional components: an \emph{input instruction} that explains the format of labeled examples, an \emph{output instruction} that specifies the regex syntax and answer format, a \emph{CoT-style output instruction}~\cite{wei2023chainofthoughtpromptingelicitsreasoning} that asks for intermediate reasoning before the final answer, and a \emph{strategy} component obtained from GEPA~\cite{agrawal2026gepareflectivepromptevolution} optimization. 

\noindent \textbf{Results.} The most important component is the \emph{output instruction} (see Appendix~\ref{appendix:prompt-ablation}), which suggests that models need a clear specification of the output language and answer format; otherwise, many failures come from producing invalid or unusable regexes rather than from the induction problem alone. 
In contrast, explicitly prompting for \emph{CoT} does not consistently improve final success. The most likely reason for the lack of improvement is because GPT-oss already separates reasoning from final answers, while many recent reasoning models perform latent reasoning internally. As a result, prompt-level \emph{CoT} guidance offers limited additional benefit once the model already uses a reasoning-style decoding interface.
The GEPA-optimized \emph{strategy} component provides some benefit, especially when regexes are easier at lower \emph{StarDepth}. However, the gain is limited and becomes less clear on harder regexes. This result may indicate mild overfitting of the optimized strategy to simpler regex structures, and does not seem to provide a general solution for more difficult tasks.

Overall, the results suggest that prompt design is relatively robust once the input and output formats, and the syntax of the desired output regex are specified clearly. The exact wording of auxiliary reasoning or strategy instructions has a smaller effect.

\subsection{RQ5: How Do Different Ways of Providing Counterexamples Affect Performance?}

\noindent \textbf{Setup.} We study whether clustered counterexamples affect learning.

\noindent \textbf{Results.} Table~\ref{tab:clustered-vs-nonclustered} shows that across both datasets, clustered counterexamples consistently improve success rates. This result suggests that the benefit of counterexamples depends not only on adding more labeled strings, but also on how those strings cover the disagreement between the predicted and target languages. Clustered counterexamples provide a more systematic view of the error region, making the revealed mismatch clearer and less redundant. As a result, they give the model a stronger signal for diagnosing the failed hypothesis and revising the regex.

\begin{table*}[t]
\centering
\small
\setlength{\tabcolsep}{8pt}
\begin{tabular}{lcccccc}
\toprule
\multirow{2}{*}{Setting}
& \multicolumn{3}{c}{Simple}
& \multicolumn{3}{c}{Extended} \\
\cmidrule(lr){2-4}
\cmidrule(lr){5-7}
& SD=1 & SD=2 & SD=3
& SD=0 & SD=1 & SD=2 \\
\midrule
Non-clustered
& 90.5\% & 47.6\% & 52.4\%
& 100.0\% & 76.2\% & 61.1\% \\
Clustered
& 100.0\% & 85.7\% & 76.2\%
& 100.0\% & 81.0\% & 77.8\% \\
\bottomrule
\end{tabular}
\caption{Comparison between non-clustered and clustered counterexamples on both datasets. Each row uses regularization and the full agentic workflow. Each \emph{StarDepth} column aggregates 21 regexes, with 3 regexes from every \emph{\#States}, one run per regex.}
\label{tab:clustered-vs-nonclustered}
\end{table*}

\subsection{Ablations}
We ablate the main components of our workflow in Table~\ref{tab:simplyrx-algorithm-components-ablation}. 
\begin{table*}[t]
\centering
\small
\setlength{\tabcolsep}{4pt}
\begin{tabular}{lcccc ccc}
\toprule
Ablation & C-CE & Reg. & Refl. & Repair & \shortstack{SD=1} & \shortstack{SD=2} & \shortstack{SD=3} \\
\midrule
Standard & $\times$ & $\checkmark$ & $\times$ & $\times$ & 52.4\% & 14.3\% & 19.0\% \\
Single Inference & $\checkmark$ & $\checkmark$ & $\times$ & $\times$ & 85.7\% & 71.4\% & 61.9\% \\
No Reg & $\checkmark$ & $\times$ & $\checkmark$ & $\checkmark$ & 100.0\% & 85.7\% & 71.4\% \\
No Reflection & $\checkmark$ & $\checkmark$ & $\times$ & $\checkmark$ & 95.2\% & 76.2\% & 76.2\% \\
No Repair-loop & $\checkmark$ & $\checkmark$ & $\checkmark$ & $\times$ & 85.7\% & 66.7\% & 66.7\% \\
Agentic & $\checkmark$ & $\checkmark$ & $\checkmark$ & $\checkmark$ & 100.0\% & 85.7\% & 76.2\% \\
\bottomrule
\end{tabular}
\caption{Algorithm-components ablation on Simply Regex Dataset. \emph{C-CE}, \emph{Reg.}, \emph{Refl.}, \emph{Repair} indicates whether clustered counterexamples, regularization, reflection, repair loops are used. Each \emph{StarDepth} column aggregates 21 runs, with 3 regexes from every \emph{\#States}, one run per regex.}
\label{tab:simplyrx-algorithm-components-ablation}
\vspace{-1em}
\end{table*}
Regularization has only a small effect on final success. From the trajectories, we observe that without regularization the model is more likely to overfit early, when only a few examples are available, by producing long and overly specific regexes. As more counterexamples are added, this issue is naturally reduced: the larger and more diverse training set makes it harder for a highly specific expression to fit all observations, and the model is pushed toward more general patterns. Consequently, removing regularization does not substantially hurt final performance. 
Reflection and repair loops both contribute to the performance, but in different ways. Empirically, repair loops provide the larger gain, which is expected because they use additional inference tokens to perform multiple rounds of regex generation and verification. The Extended Regex ablation and Pareto plots are provided in Appendix~\ref{appendix:ablation}. The Pareto plot in Figure~\ref{fig:pareto-simplyrx-all} shows the same trade-off: \textsc{Single} lies on a strong efficiency frontier, while Agentic variants usually achieve higher success, but require more tokens.
We also evaluate a multi-candidate repair variant, where $10$ regexes are proposed simultaneously each round, and the one with the best repair score is selected. This variant does not outperform \textsc{Single}, achieving \(81.0\%\), \(61.9\%\), and \(61.9\%\) success at \(\mathrm{SD}=1,2,3\), versus \(85.7\%\), \(71.4\%\), and \(61.9\%\) for \textsc{Single}. This result suggests that broader candidate exploration alone is insufficient to improve refinement.

\section{Conclusion}
Counterexample-guided feedback provides a controllable way to study how LLMs use feedback for symbolic learning. We showed that verifier-generated counterexamples---especially clustered counterexamples---substantially improve regex induction. More broadly, our findings suggest that in verifiable domains, for which informative failures can be synthesized, LLM agents can use structured feedback to move beyond merely fitting examples and toward reliable iterative refinement.

\newpage
\bibliographystyle{plainnat}
\bibliography{references}

\newpage
\appendix

\section{Background Details}
\label{appendix:background_details}

We give formal definition of membership oracle queries and equivalence oracle queries:

\noindent\paragraph{Membership Oracle Queries.}
A membership oracle labels individual instances. For a queried instance $x \in \mathcal{X}$, it returns
\[
\mathcal{O}_{\mathrm{mem}}(x)=\mathbf{1}[x \in L^\star].
\]
Membership queries provide local supervision about the behavior of the target language on specific examples. In an active setting, these examples may be selected to be informative. However, finitely many membership labels are generally insufficient to identify the full target language.

\noindent\paragraph{Equivalence Oracle Queries.}
An equivalence oracle evaluates an entire hypothesis. Given a proposed hypothesis $h$, the oracle checks whether $L(h)=L^\star$. If the two languages are equivalent, learning succeeds. Otherwise, the oracle returns counterexamples
\[
C \subseteq L(h) \triangle L^\star,
\]
where $\triangle$ denotes symmetric difference. Each counterexample exposes a concrete disagreement between the hypothesis and the target, indicating either that the hypothesis is too restrictive or that it over-generalizes.

\section{Pseudo Codes for Agentic Workflow}
\label{appendix:pseudo_code}

Algorithm~\ref{alg:outer-agentic-loop} describes the outer loop. At each epoch, the teacher checks the current hypothesis and returns counterexamples when the hypothesis is incorrect. These counterexamples are added to the aggregate training set. The learner is then prompted to reflect on the previous hypothesis, its reasoning, and the newly revealed failures before producing an updated hypothesis.

Algorithm~\ref{alg:inner-repair-loop} describes the inner repair loop. For each retry, the learner proposes a reasoning trace and a regex. The teacher evaluates whether the regex compiles, whether it fits the accumulated examples, and whether it is equivalent to the target. If the candidate fails, the resulting feedback is converted into a repair prompt for the next retry.

\section{Detailed information of Regex Datasets}
\label{appendix:comp-simple-extended}

We summarize the main syntax feature differences between Simple Regex and Extended Regex in Table~\ref{tab:simple-vs-extended-regex}.

For Simple Regex dataset, in order to synthesize regular expressions at scale, we recursively sample syntax trees over union, concatenation, and Kleene-star under size and star-depth budgets. We use randomness to efficiently cover a diverse range of structures without exhaustively enumerating the search space. Afterwards, we simplify duplicated Kleene-stars, for instance, reducing ``$(a*)*$'' to ``$a*$'', and then construct the actual regular expressions out of those syntax trees.

The extended regexes used in our experiments come from the KB13 dataset~\cite{kushman-barzilay-2013-using}. Because the expressions from KB13 have a slightly different syntax, we first parse and then covert them into our desired form. We then group these regexes by the complexity criteria described in Section~\ref{experiments}, which guides the selection of our final Extended dataset.

\begin{table}[t]
\centering
\small
\begin{tabular}{p{0.23\linewidth} p{0.34\linewidth} p{0.34\linewidth}}
\hline
\textbf{Feature} & \textbf{Simple Regex} & \textbf{Extended Regex} \\
\hline
Alphabet & $\Sigma=\{a,b,c\}$ & $\Sigma = [A\text{-}Za\text{-}z0\text{-}9\#]$ \\

Literal symbols & Single characters or literal \texttt{epsilon} & Any single symbol in $\Sigma$ \\

Concatenation & Space-separated symbols, e.g., \texttt{a b c} & Implicit by adjacency, e.g., \texttt{abc} \\

Union & \texttt{+} & \texttt{|} \\

Kleene star & \texttt{*} & \texttt{*} \\

One-or-more & Not supported & \texttt{+} \\

Optional & Not supported & \texttt{?} \\

Counted repetition & Not supported & \texttt{\{n\}}, \texttt{\{n,m\}}, \texttt{\{n,\}} \\

Grouping & Parentheses allowed for scope and precedence & Parentheses allowed for scope and precedence \\

Character classes & Not supported & Allowed, e.g., \texttt{[A-Z]}, \texttt{[0-9]}\\

Intersection & Not supported & \texttt{\&} \\

Complement / negation & Not supported & $\sim$\texttt{R} \\

Epsilon & Explicit literal \texttt{epsilon} & No dedicated literal emphasized \\

Precedence / associativity & Not explicitly specified beyond using parentheses for control & Priority: quantifiers $>$ $\sim$ $>$ concatenation $>$ \texttt{\&} $>$ \texttt{|}; concatenation, \texttt{\&}, and \texttt{|} are left-associative \\

Typical example & \texttt{(a b + a c)*} & \texttt{(ab|ac)*} \\
\hline
\end{tabular}
\caption{Comparison of the syntax features for Simple Regex and Extended Regex}
\label{tab:simple-vs-extended-regex}
\end{table}

\section{Detailed Configuration and Setup}
\label{appendix:setup-details}

The learner is implemented by LLMs. We use the open-source model ``openai/gpt-oss-120b''~\cite{openai2025gptoss120bgptoss20bmodel} (also ``GPT''~\cite{OpenAI2026GPT54} and ``Qwen''~\cite{qwen3technicalreport} models for model ablation). For the main experiments, we perform three trials 
per regex for robustness and reliability. And from RQ3 to RQ5 in Section~\ref{experiments}, we perform only one trial per regex for relevant experiments. For all the experiments, we set the maximum context-window length to be $65536$, the maximum output-token length to be $32768$ (including both potential reasoning tokens and revealed output tokens), and the temperature to be $0.0$.

For ``openai/gpt-oss-120b'' experiments, we use 2 NVIDIA L40S GPUs to load the model and do inference. The CPU memory used is less than $64$GB, and disk usage is less than $200$GB. For other models (``GPT-5.4'' and ``Qwen3-235B''), we use API calls for inference.

For implementation of regex, we use the off-the-shelf python library ``pyformlang''~\cite{pyformlang}. The maximum length of the training examples is set to be $32$. For \textsc{Standard} setting, we start with $3$ training examples, and scale logarithmically up to maximally $3000$ before reporting failure. For settings involving counterexamples, we maximally starts with $8$ initial labeled examples, and run maximally $12$ epochs with number of counterexamples added each round clipped to $250$ maximally. For settings involving repair loops, the repair budget is set to be $3$.

The main metric that we use to evaluate the performance is run-level success ratio. For example, for the main results in Table~\ref{tab:main-result-success-by-stardepth}, we evaluate 21 regexes per cell, and run 3 times per regex, yielding 63 runs in total, and we report the ratio of the number of successful runs over the number of total runs.

\section{More Results}

\subsection{Heatmaps of Main Evaluation}
\label{appendix:heatmap-main}

\begin{figure}[ht]
    \centering
    \includegraphics[width=\linewidth]{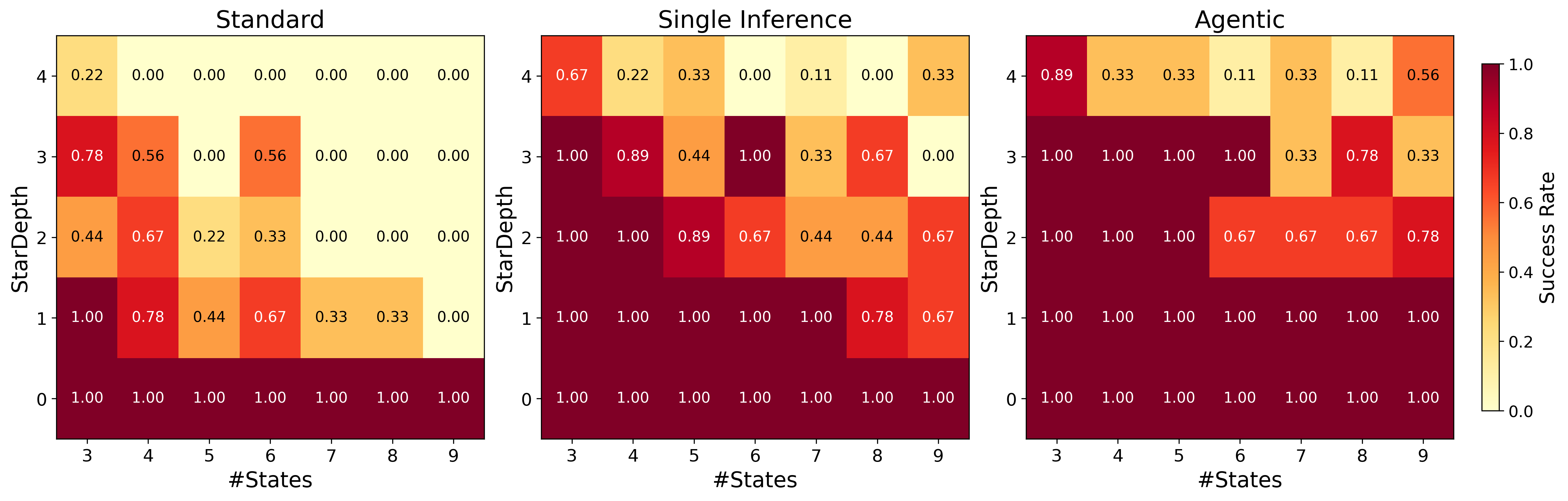}
    \caption{Success rate heatmaps across regex complexity for main Simple Regex experiments. Rows correspond to \emph{StarDepth} and columns correspond to the number of DFA states. Darker colors indicate higher success rates.}
    \label{fig:heatmap-simplyrx}
\end{figure}

\begin{figure}[ht]
    \centering
    \includegraphics[width=\linewidth]{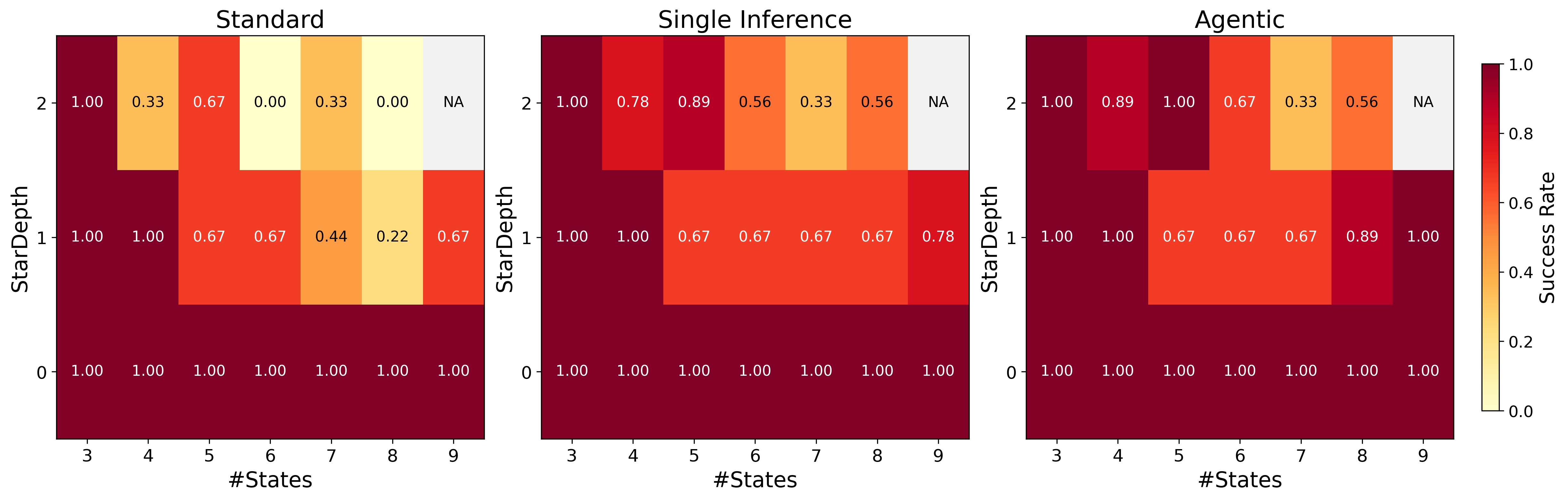}
    \caption{Success rate heatmaps across regex complexity for main Extended Regex experiments. Rows correspond to \emph{StarDepth} and columns correspond to the number of DFA states. Darker colors indicate higher success rates.}
    \label{fig:heatmap-extrx}
\end{figure}

Figures~\ref{fig:heatmap-simplyrx} and~\ref{fig:heatmap-extrx} show the full main-evaluation heatmaps over both \emph{StarDepth} and the number of DFA states. These plots complement the aggregated results in Table~\ref{tab:main-result-success-by-stardepth}, where we report success rates grouped by \emph{StarDepth}.

The heatmaps show that \emph{StarDepth} is a stronger indicator of learning difficulty than the number of states. Within the same \emph{StarDepth} level, performance varies with the number of states, but the overall degradation is more clearly aligned with increasing \emph{StarDepth}. This motivates our choice to aggregate the main results by \emph{StarDepth} in the main text.

The heatmaps also confirm the main trend: counterexample-guided settings improve success rates across a broad range of complexity levels. The improvement is most visible in higher-\emph{StarDepth} regions, where \textsc{Standard} prompting often fails but counterexample-guided methods recover a larger fraction of target regexes.

\subsection{Sample Efficiency Comparison}
\label{appendix:sample-eff}

\begin{figure}[ht]
    \centering
    \includegraphics[width=\linewidth]{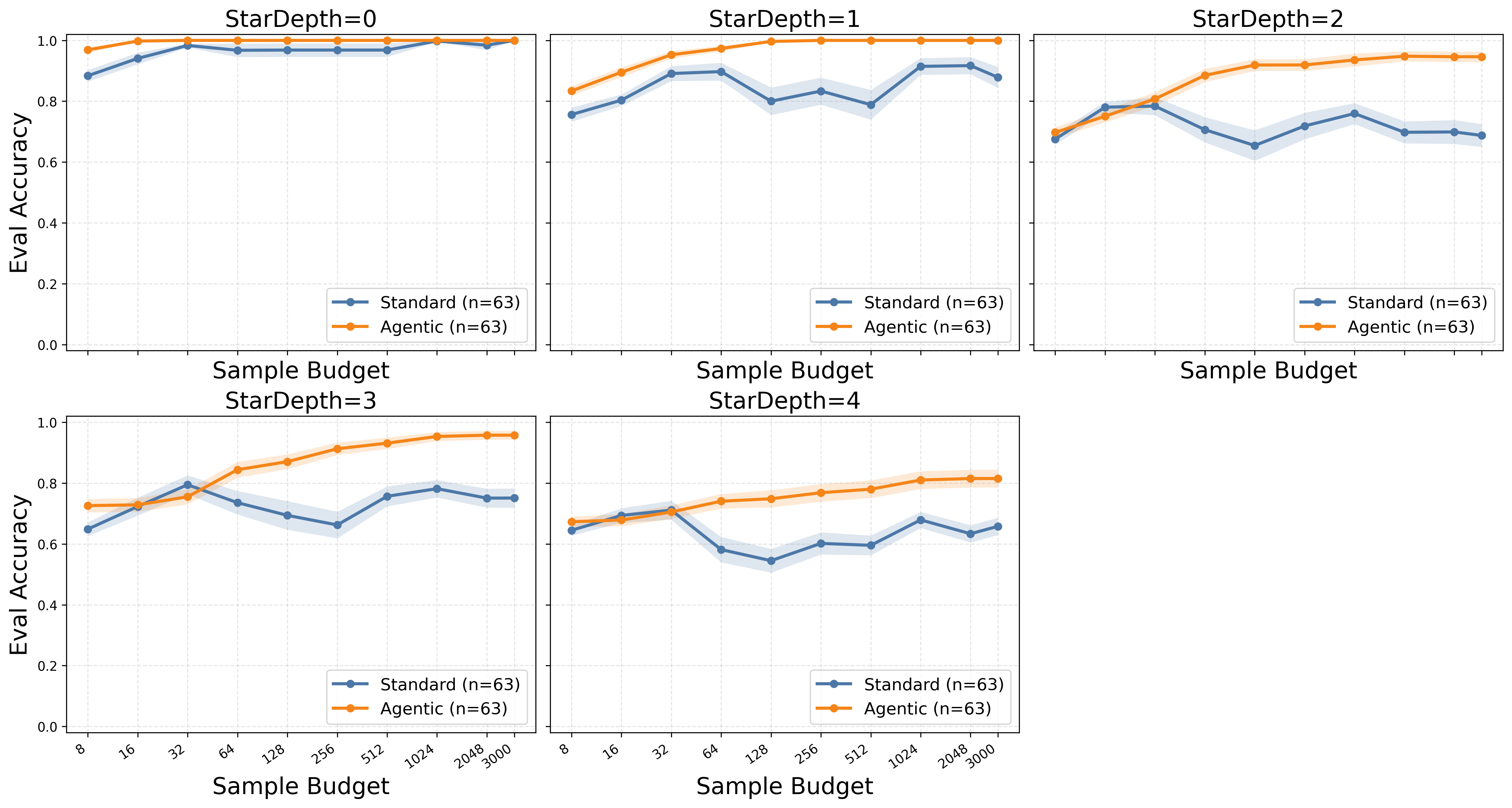}
    \caption{Evaluation accuracy of hypothesis selected under sample budget on each \emph{StarDepth} of Simple Regex Dataset. For each budget, we take the last hypothesis in the learning trajectory whose cumulative number of labeled training examples does not exceed that budget, and evaluate it on the corresponding evaluation set. Each point is the mean evaluation accuracy, and the shaded region indicates variation.}
    \label{fig:sample-eff-simplyrx}
\end{figure}

\begin{figure}[ht]
    \centering
    \includegraphics[width=\linewidth]{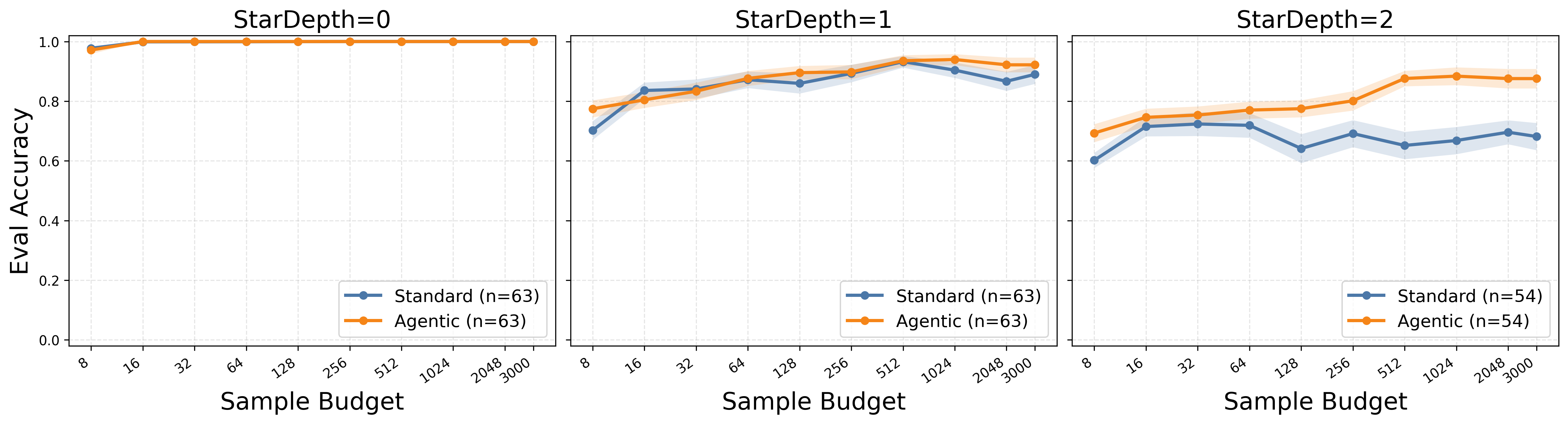}
    \caption{Evaluation accuracy of hypothesis selected under sample budget on each \emph{StarDepth} of Extended Regex Dataset. For each budget, we take the last hypothesis in the learning trajectory whose cumulative number of labeled training examples does not exceed that budget, and evaluate it on the corresponding evaluation set. Each point is the mean evaluation accuracy, and the shaded region indicates variation.}
    \label{fig:sample-eff-extrx}
\end{figure}

Figures~\ref{fig:sample-eff-simplyrx} and~\ref{fig:sample-eff-extrx} compare \textsc{Standard} and \textsc{Agentic} as a function of sample budget. For each budget, we take the last hypothesis in the learning trajectory whose cumulative number of labeled training examples does not exceed that budget, and evaluate this hypothesis on the corresponding evaluation set, which contains sufficiently $800$ labeled strings. The curves report mean evaluation accuracy across all regexes in the corresponding \emph{StarDepth} group, and the shaded regions indicate variation across regexes.

Across both datasets, \textsc{Agentic} generally reaches higher evaluation accuracy at the same sample budget, and it tends to improve more steadily as more examples are observed. The advantage is most visible at nontrivial difficulty levels, where \textsc{Standard} often plateaus earlier or fluctuates more, while \textsc{Agentic} continues to refine stronger hypotheses. This pattern indicates that verifier-guided counterexamples help the learner extract more information from each additional example.

Taken together, these curves show that our method is more sample-efficient: it typically achieves better intermediate hypotheses with fewer labeled examples, and this advantage persists across a broad range of budgets and difficulty levels.

\subsection{Details on Robustness of Prompt Design}
\label{appendix:prompt-ablation}

Table~\ref{tab:simplyrx-prompt-ablation} reports a prompt ablation for the \textsc{Single} setting on the Simple Regex dataset. The largest improvement comes from adding output-side instructions (syntax and output format requirements): compared with naive prompting, specifying the regex syntax and answer format raises success from $14.3\%/0.0\%/4.8\%$ to $71.4\%/33.3\%/23.8\%$ across \emph{StarDepth} 1--3. Adding input-format instructions provides smaller gains, while explicit \emph{CoT} prompting and the GEPA-optimized \emph{strategy} yield only modest additional improvements. These results suggest that prompt design is relatively robust once the output structure is clearly specified.

\begin{table*}[t]
\centering
\small
\setlength{\tabcolsep}{4pt}
\begin{tabular}{llccccccc}
\toprule
Method & Prompt & Input & Output & CoT & Strategy & \shortstack{SD=1} & \shortstack{SD=2} & \shortstack{SD=3} \\
\midrule
\multirow{4}{*}{\shortstack{Single}} & Naive & $\times$ & $\times$ & $\times$ & $\times$ & 14.3\% & 0.0\% & 4.8\% \\
 & Input & $\checkmark$ & $\times$ & $\times$ & $\times$ & 19.0\% & 0.0\% & 9.5\% \\
 & Output & $\times$ & $\checkmark$ & $\times$ & $\times$ & 71.4\% & 33.3\% & 23.8\% \\
 & IO & $\checkmark$ & $\checkmark$ & $\times$ & $\times$ & 71.4\% & 38.1\% & 33.3\% \\
 & Zero & $\checkmark$ & $\checkmark$ & $\checkmark$ & $\times$ & 71.4\% & 33.3\% & 33.3\% \\
 & Full & $\checkmark$ & $\checkmark$ & $\checkmark$ & $\checkmark$ & 85.7\% & 38.1\% & 38.1\% \\
\bottomrule
\end{tabular}
\caption{Prompt ablation on Simply Regex varying included prompt components. For simplicity, we directly use non-clustered counterexamples without regularization and without agentic refinements. Each \emph{StarDepth} column aggregates 21 runs, with 3 regexes from every \emph{\#States}, one run per regex.}
\label{tab:simplyrx-prompt-ablation}
\end{table*}

\subsection{More Ablation Study}
\label{appendix:ablation}

This section provides additional ablation results for the algorithmic components. Table~\ref{tab:extrx-algorithm-components-ablation} reports the component ablation on the Extended Regex dataset, complementing the Simple Regex ablation in the main text. The Extended Regex results are broadly consistent with the main findings. Using counterexamples improves over the \textsc{Standard} setting, and the largest gains appear on harder tasks. The agentic variants further improve performance relative to using counterexamples only as additional training data, although the effect of each individual component is not strictly monotonic. In particular, removing regularization or reflection can sometimes achieve comparable or higher success on this dataset, suggesting that these components affect the balance between exploration, correction, and overfitting rather than acting as uniformly beneficial switches.

\begin{table*}[ht]
\centering
\small
\setlength{\tabcolsep}{4pt}
\begin{tabular}{lcccc ccc}
\toprule
Ablation & C-CE & Reg. & Refl. & Repair & \shortstack{SD=0} & \shortstack{SD=1} & \shortstack{SD=2} \\
\midrule
Standard & $\times$ & $\checkmark$ & $\times$ & $\times$ & 100.0\% & 66.7\% & 38.9\% \\
Single Inference & $\checkmark$ & $\checkmark$ & $\times$ & $\times$ & 100.0\% & 71.4\% & 55.6\% \\
No Reg & $\checkmark$ & $\times$ & $\checkmark$ & $\checkmark$ & 100.0\% & 76.2\% & 83.3\% \\
No Reflection & $\checkmark$ & $\checkmark$ & $\times$ & $\checkmark$ & 100.0\% & 85.7\% & 88.9\% \\
No Repair-loop & $\checkmark$ & $\checkmark$ & $\checkmark$ & $\times$ & 100.0\% & 76.2\% & 77.8\% \\
Agentic & $\checkmark$ & $\checkmark$ & $\checkmark$ & $\checkmark$ & 100.0\% & 81.0\% & 77.8\% \\
\bottomrule
\end{tabular}
\caption{Algorithm-components ablation on Extended Regex Dataset. \emph{C-CE}, \emph{Reg.}, \emph{Refl.}, \emph{Repair} indicates whether clustered counterexamples, regularization, reflection, repair loops are used. Each \emph{StarDepth} column aggregates 21 runs, with 3 regexes from every \emph{\#States}, one run per regex.}
\label{tab:extrx-algorithm-components-ablation}
\end{table*}

\begin{figure}[ht]
    \centering

    \begin{subfigure}[t]{0.48\textwidth}
        \centering
        \includegraphics[width=\linewidth]{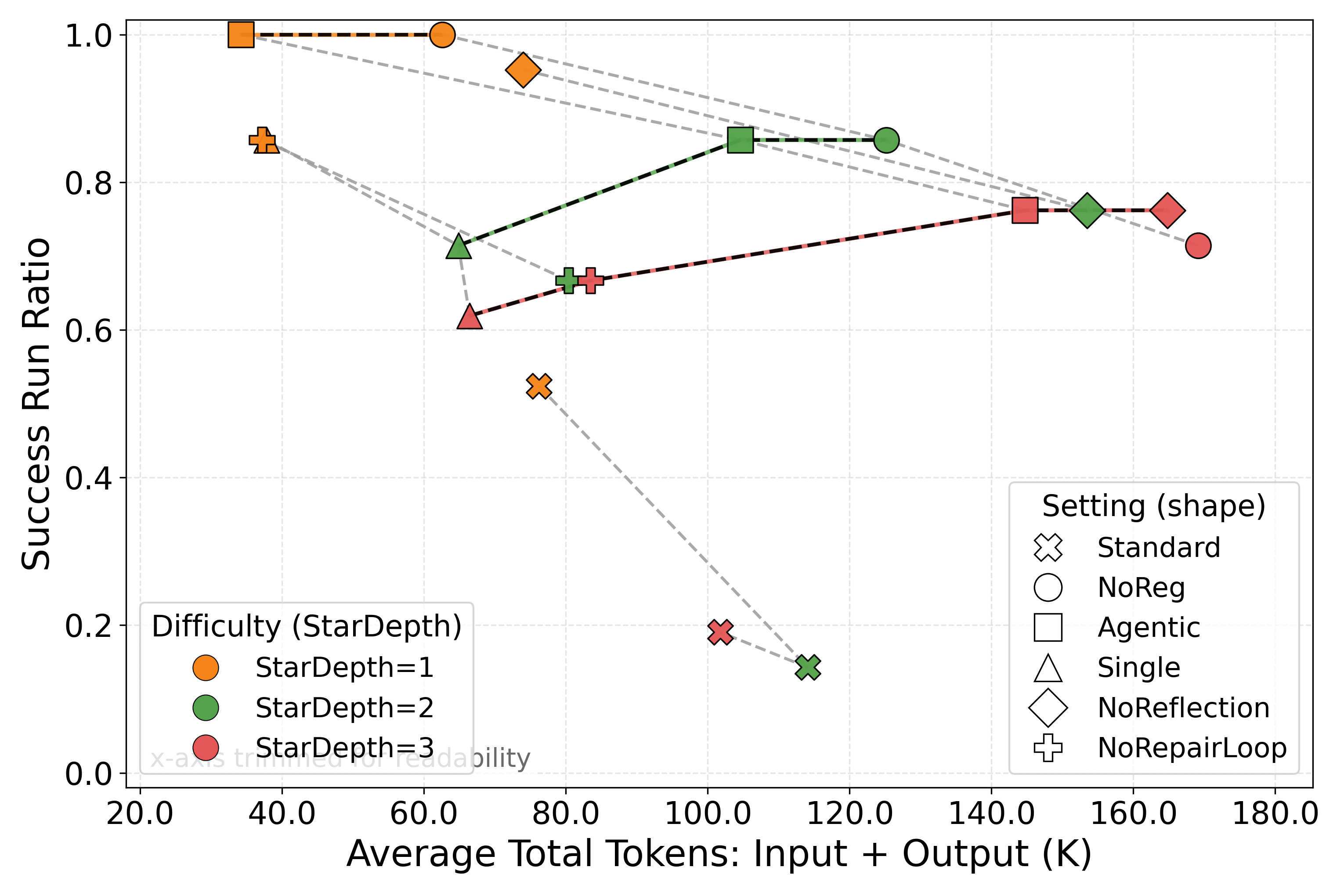}
        \caption{Component Ablation on Simple Regex Dataset}
        \label{fig:pareto-simplyrx-all}
    \end{subfigure}
    \hfill
    \begin{subfigure}[t]{0.48\textwidth}
        \centering
        \includegraphics[width=\linewidth]{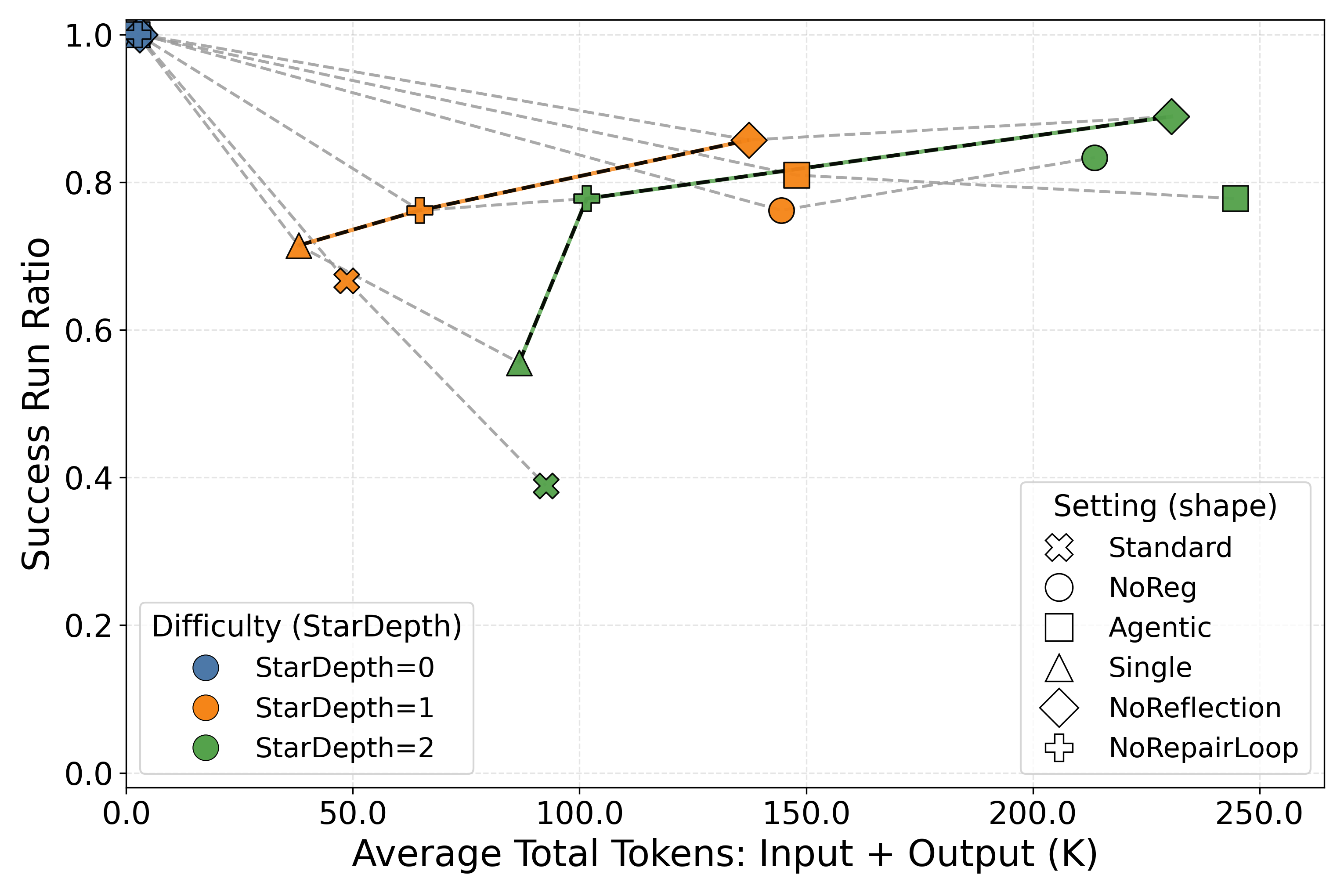}
        \caption{Component Ablation on Extended Regex Dataset}
        \label{fig:pareto-extrx-all}
    \end{subfigure}

    \caption{Pareto frontiers across full ablation settings on Simple and Extended Regex Datasets.}
    \label{fig:pareto-frontiers-all}
\end{figure}

Figure~\ref{fig:pareto-frontiers-all} shows the corresponding Pareto frontiers over success rate and total token cost for both Simple and Extended Regex. The plots illustrate the main trade-off across ablation settings: simpler counterexample-guided settings are more token-efficient, while repair-based agentic variants often achieve higher success at a larger inference cost. Overall, these results support the conclusion that counterexamples are the primary driver of improvement, while the remaining workflow components mainly control how effectively and how expensively the model uses this feedback.

\section{Prompt Templates}
\label{appendix:prompts}

This appendix documents the prompt templates used in our experiments.  Across all settings, the learner is prompted with a task description, a description of the input format, domain-specific output-syntax requirements, and an output-format instruction. Additional modules are optionally appended for regularization, clustered counterexample interpretation, reflection, repair, and multi-candidate repair.

\paragraph{Prompt Families.}
We use separate base prompt families for \textsc{Simple Regex} and \textsc{Extended Regex}. Both ask the model to infer a single regex consistent with labeled examples, but they differ in the target syntax. The \textsc{Simple Regex} prompt requires output in \texttt{pyformlang.regular\_expression.Regex} syntax over a small alphabet, while the \textsc{Extended Regex} prompt uses an extended regex syntax that includes conjunction, negation, and additional quantifiers over the fixed alphabet \(\Sigma=\texttt{[A-Za-z0-9\#]}\).

\paragraph{Regularization and Clustered Counterexamples.}
When regularization is enabled, we prepend a constraints block encouraging concise, general regexes. For \textsc{Simple Regex}, this constrains the total regex length and Kleene-star depth; for \textsc{Extended Regex}, the length bound is larger and the allowed star nesting is smaller. When clustered counterexamples are used, the input-format section is augmented with an explanation of grouped character classes, since some counterexamples are presented as character-class abstractions rather than only literal strings.

\paragraph{Agentic Update Prompts.}
In agentic settings, the base prompt is augmented with additional instructions. Reflection prompts summarize the previous reasoning, previous regex, and new counterexamples, and ask the model to explain what failed and what should change. Repair prompts ask the model to revise the previous regex using failure feedback and misclassified examples while preserving correct structure from the earlier attempt. In the multi-candidate repair ablation, the model is instead asked to generate several repaired regexes, each wrapped in its own \texttt{<ans>} block, after which the system selects the best one externally.

\paragraph{Training-Data Formatting.}
In all settings, the labeled examples are appended under a fixed header:
\begin{quote}
\texttt{Training Data (Each line has one input-output pair separated by comma):}
\end{quote}
Each subsequent line has the form \texttt{<string>, <label>}. Empty strings are represented by leaving the string field blank before the comma.

\paragraph{Base Prompt: Simple Regex.}
\begin{verbatim}
TASK
You will be given labeled strings and must infer a single regular language
that matches all positives (label 1) and rejects all negatives (label 0).
Output a concise regex in pyformlang.regular_expression.Regex syntax.

INPUT FORMAT
- You receive a block titled “Training Data (Each line has one input-output
  pair separated by comma):”.
- Each line is "<string>, <label>" where label in {1, 0}. The string may be
  empty; an empty string appears as nothing before the comma (", 1") and
  represents epsilon.
- The alphabet is exactly the set of characters appearing in the data
  (typically a, b, c). Do not introduce other symbols.
{clustered_ce_instr}

PYFORMLANG REGEX SYNTAX
- Union: +
- Concatenation: space-separated symbols.
- Kleene star: *
- Parentheses are allowed for grouping.
- Use the literal epsilon when needed.
- Do NOT use: | . ? [] {} anchors/lookarounds, multi-character tokens,
  or any symbol not present in the training data.

INFERENCE STRATEGY
1) Check start/end constraints.
2) Look for block structure and repetition.
3) Choose between star-of-union and union-of-stars.
4) Prefer compact factorizations.
5) Handle epsilon only when supported.
6) Avoid over-generalization.
7) Verify all positives and negatives before finalizing.

{regularization_instr}
{agentic_reflection_instr}

OUTPUT FORMAT
- First, provide 1-3 concise sentences wrapped in <reasoning>...</reasoning>.
- Then output ONLY the final regex wrapped in <ans>...</ans>.

Training Data (Each line has one input-output pair separated by comma):
{0}
\end{verbatim}

\paragraph{Base Prompt: Extended Regex.}
\begin{verbatim}
TASK
You will be given labeled strings and must infer a single regular language
that matches all positives (label 1) and rejects all negatives (label 0).
Output a concise regex in our specified syntax (extended from 
pyformlang.regular_expression.PythonRegex).

INPUT FORMAT
- You receive a block titled “Training Data (Each line has one input-output
  pair separated by comma):”.
- Each line is "<string>, <label>" where label in {1, 0}. The string may be
  empty; an empty string appears as nothing before the comma (", 1") and
  represents epsilon.
- The alphabet is fixed. Do not introduce other symbols.
{clustered_ce_instr}

EXT REGEX SYNTAX (Extended PythonRegex)
- Alphabet is fixed: Sigma = {sigma}
- Concatenation is implicit by adjacency.
- Union: |
- Grouping: ( ... )
- Conjunction / intersection: &
- Negation / complement: ~( ... )
- Quantifiers: *, +, ?, {n}, {n,m}, {n,}
- Do not use dot, negated character classes, anchors, word boundaries,
  lookarounds, or backreferences.

INFERENCE STRATEGY
1) Check prefix and suffix constraints.
2) Look for repeated substrings and block structure.
3) Choose between star-of-union and union-of-stars.
4) Factor common prefixes/suffixes and use character classes when appropriate.
5) Handle epsilon only when required.
6) Avoid over-generalization.
7) Verify positives, negatives, and boundary cases.

{regularization_instr}
{agentic_reflection_instr}

OUTPUT FORMAT
- First, provide 1-3 concise sentences wrapped in <reasoning>...</reasoning>.
- Then output ONLY the final regex wrapped in <ans>...</ans>.

Training Data (Each line has one input-output pair separated by comma):
{0}
\end{verbatim}

\paragraph{Regularization Blocks.}
\begin{verbatim}
Simple Regex regularization:
CONSTRAINTS
- Prefer simpler, more general regexes while staying consistent with all
  datapoints.
- Total regex length (ignoring spaces) must be <= 60 characters.
- Nesting depth of Kleene stars must be <= 4.
- Use only symbols that appear in the training data.

Extended Regex regularization:
CONSTRAINTS
- Prefer simpler, more general regexes while staying consistent with all
  datapoints.
- Total regex length (ignoring spaces) must be <= 150 characters.
- Nesting depth of Kleene stars (*, +, ?) must be <= 2.
- Use only symbols that appear in the alphabet (except metacharacters).
\end{verbatim}

\paragraph{Reflection Prompt.}
\begin{verbatim}
AGENTIC REFLECTION UPDATE
- You will receive the previous attempt's reasoning and regex, plus new
  counterexamples.
- First, briefly revise the previous reasoning to explain what failed and
  what should be changed.
- Then output an updated regex consistent with all training data and the
  counterexamples.
- Keep reasoning concise (1-3 sentences) and directly tied to the regex
  revision.

Reasoning of previous epoch:
<previous reasoning>
Regex of previous epoch:
<previous regex>
What failed in the previous epoch:
<feedback note, if available>
New counterexamples (string, label):
<counterexample lines>
\end{verbatim}

\paragraph{Repair Prompt.}
\begin{verbatim}
AGENTIC REPAIR UPDATE
- You are repairing the previous attempt using the failure feedback and
  repair examples below.
- Repair goals:
  1) Produce a regex that compiles under the required regex syntax.
  2) Fix the specific mistakes exposed by the counterexamples, disagreement
     witness, and any reported errors.
  3) Preserve the parts of the previous solution that still fit the
     training data.
- If the previous regex is invalid, first correct the syntax.
- If a witness or repair example is handled incorrectly, revise the regex
  so that string gets the correct label.
- Keep the reasoning concise and focused on what changed.
- Return the repaired regex in the required output format.

Reasoning of previous attempt:
<previous reasoning>
Regex of previous attempt:
<previous regex>
Repair feedback:
<feedback note, if available>
Repair examples (string, label):
<repair-example lines>
\end{verbatim}

\paragraph{Multi-Candidate Repair Prompt.}
\begin{verbatim}
CANDIDATE REGEX GENERATION INSTRUCTION
- Do not return only one repaired regex. Generate exactly K diverse
  candidate regexes.
- Wrap every candidate regex in its own <ans> and </ans> block.
- You may include one concise shared <reasoning>...</reasoning> block
  before the candidates.
- Do not include any text after the final </ans> block.
\end{verbatim}

\paragraph{Direct Output Format.}
\begin{verbatim}
OUTPUT FORMAT
- Output ONLY the final regex wrapped in <ans> and </ans>.
\end{verbatim}

\newpage
\begin{algorithm}[t]
\caption{Outer Agentic Learning Loop}
\label{alg:outer-agentic-loop}
\begin{algorithmic}[1]
\Require Initial labeled examples $\mathcal{D}_0$, task instructions $I$, reflection instructions $I_{\mathit{refl}}$, teacher $T$, learner $M$, maximum epochs $E$, repair budget $R$
\Ensure Final hypothesis regex $\hat r$

\State Initialize aggregate training set $\mathcal{C} \gets \emptyset$
\State Initialize current best hypothesis $\hat r \gets \textsc{None}$
\State Initialize current best reasoning $\hat z \gets \textsc{None}$

\For{$e = 1,\ldots,E$}

    \If{$e = 1$}
        \State $c \gets \mathcal{D}_0$
        \State $p_{\mathit{refl}} \gets \emptyset$
    \Else
        \State Teacher $T$ checks $\hat r$ against the target language
        \State Gather counterexamples from Teacher $T$ feedback:
        \[
        c=\{(x,y): \hat r(x) \neq y\}\subseteq L(r)\triangle L^\star
        \]
        \State Build reflection prompt $p_{\mathit{refl}} \gets \textsc{BuildReflectionPrompt}(I_{\mathit{refl}}, \hat r, z, c)$
    \EndIf

    \State Update aggregate training set: $\mathcal{C} \gets \mathcal{C} \cup c$
    \State Construct base prompt $p_e$ from: $(I, \mathcal{C}, p_{\mathit{refl}})$

    \State Run inner repair loop:
    \[
    (r, z, a) \gets \textsc{RepairLoop}(p_e, \mathcal{C}, T, M, R)
    \]

    \State Update current best hypothesis: $\hat r \gets r$, $\hat z \gets z$

    \If{$a = 1$}
        \State \Return $\hat r$
    \EndIf

\EndFor

\State \Return $\hat r_{\mathrm{prev}}$
\end{algorithmic}
\end{algorithm}

\begin{algorithm}[t]
\caption{Inner Reflection--Repair Loop}
\label{alg:inner-repair-loop}
\begin{algorithmic}[1]
\Require Epoch prompt $p_e$, examples $\mathcal{C}$, repair instructions $I_{\mathit{rep}}$, teacher $T$, learner $M$, retry budget $R$
\Ensure Best regex $\hat r$, best reasoning $\hat z$, solved flag $a$

\State Initialize the repair prompt, $p_{\mathit{rep}} \gets \emptyset$.
\State Initialize the best score, $\mathrm{bestScore} \gets -\infty$.
\State Initialize the best candidate, $(\hat r, \hat z) \gets (\textsc{None}, \textsc{None})$.

\For{$t = 1,\ldots,R$}
    \State Query the LLM-learner: $(r, z) \gets M(p_e \oplus p_{\mathit{rep}})$

    \State Syntax error message from compilation $m \gets \text{compile}(r)$
    \State Equivalence flag from equivalence queries: $a \gets T.\mathrm{Equiv}(r)$
    \State Gather misclassified exmaples $e\gets \{(x,y)\in\mathcal{C}:r(x)\neq y\}$

    \If{$a = 1$}
        \State $\Return(r,z, 1)$
    \EndIf

    \State $\mathrm{score} \gets \textsc{RepairScore}(c,e,\mathcal{C})$

    \If{$\mathrm{score} > \mathrm{bestScore}$}
        \State $\mathrm{bestScore} \gets \mathrm{score}$
        \State $(\hat r, \hat z) \gets (r, z)$
    \EndIf

    \State Build the next repair prompt 
    \[p_{\mathit{rep}} \gets \textsc{BuildRepairPrompt}(I_{\mathit{rep}}, r,z,m,e)\]
\EndFor

\State $\Return(\hat r, \hat z, 0)$

\end{algorithmic}
\end{algorithm}

\end{document}